\documentclass[conference]{IEEEtran}
\IEEEoverridecommandlockouts
\usepackage[T1]{fontenc}
\usepackage{multirow}
\usepackage{tabularx,array}
\usepackage{makecell}
\usepackage{url}
\renewcommand{\url}[1]{}

\usepackage{cite}
\usepackage{amsmath,amssymb,amsfonts}
\usepackage{algorithmic}
\usepackage{graphicx}
\usepackage{textcomp}
\usepackage{xcolor}

\def\BibTeX{{\rm B\kern-.05em{\sc i\kern-.025em b}\kern-.08em
    T\kern-.1667em\lower.7ex\hbox{E}\kern-.125emX}}
\begin{document}

\title{Threats to Arabic Handwriting Recognition: Investigating Black-Box Adversarial Attacks on embedded ConvNet models\\
}

\author{\IEEEauthorblockN{1\textsuperscript{st} Mohsine El Khayati}
\IEEEauthorblockA{\textit{Systems theory and informatics laboratory} \\
\textit{Moulay Ismail University of Meknes}\\
Meknes, Morocco \\
m.elkhayati@umi.ac.ma}

\and
\IEEEauthorblockN{2\textsuperscript{nd} Abdelillah Semma}
\IEEEauthorblockA{\textit{Department of Computer Science} \\
 \textit{EST of Sidi Bennour}\\
\textit{Chouaib Doukkali University}\\
Eljadida, Morocco \\
semma\_abdelillah@yahoo.fr}
\and
\IEEEauthorblockN{3\textsuperscript{rd} Abdelaziz Courr}
\IEEEauthorblockA{\textit{Faculty of Education Sciences} \\
\textit{University Mohammed V}\\
Rabat, Morocco \\
abdelaaziz.courr@gmail.com}
\and
\IEEEauthorblockN{4\textsuperscript{th} Rachid Elouahbi}
\IEEEauthorblockA{\textit{Laboratory of Computer Science and Applications} \\
\textit{Moulay Ismail University of Meknes}\\
Meknes, Morocco \\
r.elouahbi@umi.ac.ma}
}

\maketitle

\begin{abstract}
Arabic handwriting recognition (AHR) has made significant progress with deep learning models. AHR research has largely focused on performance, with security receiving little attention. This study provides what appears to be a new line of inquiry by demonstrating the vulnerability of high-performing models to adversarial black-box attacks. The focus on black-box attacks reflects real-world scenarios where the attacker has no prior knowledge of the model architecture. Extensive experiments were conducted on two benchmark AHR datasets containing Arabic handwritten Characters. Results demonstrated the effectiveness of the attacks, with the Pixle attack achieving an attack success rate of 99-100\% on most models. Other, less aggressive attacks achieved success rates of 50-96\% across most experiments. Despite the higher attack success rate, the attacks maintain the structural integrity of the characters, rendering them almost imperceptible to the human eye. The findings indicate the higher vulnerability of the studied models to adversarial manipulation. This underscores the need to strengthen efforts to secure these models and ensure their reliability in AHR real-world applications.
\end{abstract}


\section{Introduction}

Arabic Handwriting Recognition (AHR) is an important technology with applications in document digitization, historical manuscript preservation, automated data entry in banking, education, and government sectors, as well as in recording information in day-to-day life \cite{tagougui_online_2012}. Its development enables efficient retrieval and processing of Arabic textual data, facilitating broader access to information and reducing manual transcription efforts. However, Arabic script poses significant challenges to recognition models due to its cursive nature, the visual similarities between characters, and the presence of diacritical marks, etc. \cite{korichi_arabic_2020,rabi_convolutional_2024}.

Recent advancements in deep learning, particularly through Convolutional Neural Networks (ConvNets), have significantly propelled progress in AHR. ConvNet-based architectures have proven highly effective in capturing the intricate spatial characteristics of Arabic script \cite{el_khayati_cnn-based_2024, rabi_convolutional_2024, shtaiwi_end--end_2022, ahmed_novel_2021, elleuch_novel_2016,ElKhayati2025_laveraging}. While ConvNets have proven to be effective, they remain vulnerable to adversarial attacks \cite{huang_survey_2020}. Adversarial attacks exploit model weaknesses by introducing subtle, often imperceptible perturbations to input images, leading to misclassifications even using well-trained networks \cite{akhtar_threat_2018}, while the images appear natural to human observers. Such adversarial samples can mislead models with high confidence, as demonstrated by the Fast Adversarial Watermark Attack (FAWA), which achieves a 100\% attack success rate by embedding perturbations as watermarks \cite{chen_fawa_2021}.

Adversarial attacks on AHR models can pose substantial risks, as they can compromise the accuracy and integrity of digitized documents, potentially leading to misinformation or data manipulation. This vulnerability is especially critical in sectors that rely on accurate document digitization, such as the energy and mining industries \cite{castro_improvement_2021}, automatic grading of handwritten exam responses \cite{devan_one-word_2020}, recognizing religious texts \cite{mohd_quranic_2021}, and banking.

Despite the increasing attention to adversarial attacks in various deep learning domains, research in AHR has predominantly centered on enhancing recognition performance, with a lack of focus on adversarial robustness and security issues. To the best of our knowledge, no prior studies have systematically examined the security of AHR models against adversarial threats, leaving a critical gap in the literature that warrants urgent investigation.

This study seeks to bridge the existing gap by initiating a focused investigation into the impact of black-box adversarial attacks on AHR. In particular, we assess the vulnerability of ConvNet-based embedded models trained for Arabic Handwritten Character Recognition (AHCR) under realistic attack scenarios, where the attacker has no access to model architecture or parameters. By systematically evaluating multiple lightweight architectures, this work not only uncovers the extent to which current AHR models are exposed to adversarial manipulation but also lays the groundwork for future research aimed at developing robust defense mechanisms tailored to the unique challenges of AHR.

The remainder of this paper is organized as follows: Section 2 discusses related works. Section 3 outlines the proposed methodology. Section 4 discusses the results and key findings, and Section 5 concludes with suggestions for future research directions.

\section{Related works}

Adversarial attacks have become a major focus in deep learning research, especially since Szegedy et al. \cite{szegedy_intriguing_2014} first demonstrated how small, imperceptible changes to inputs could lead to significant misclassifications in neural networks. These vulnerabilities arise due to the complexity and non-linearity of deep learning models, which make them sensitive to even minimal perturbations. These attacks target the model’s learned representations, revealing critical weaknesses in how neural networks process information \cite{akhtar_threat_2018}. Adversarial attacks introduce perturbations - small, often imperceptible changes - added to a clean image, creating an adversarial example designed to mislead a model into making incorrect predictions \cite{szegedy_intriguing_2014}.

Adversarial attacks can be categorized based on the attacker's knowledge of the model. White-box attacks, like FGSM \cite{goodfellow_explaining_2015} and PGD \cite{madry_towards_2019}, operate with complete access to the model’s architecture and parameters, allowing precise perturbations that maximize the model’s error \cite{irfan_towards_2021}. In contrast, black-box attacks operate with limited or no knowledge of the model’s internal structure, relying instead on querying outputs. Attacks like Pixle \cite{pomponi_pixle_2022} and Square \cite{andriushchenko_square_2020} fall into this category and are especially relevant for real-world applications where models are typically deployed as black-box systems \cite{agrawal_impact_2021,sen_adversarial_2023}. These methods require fewer resources than white-box attacks and can be highly effective even with minimal information \cite{huang_adversarial_2017}.

Adversarial attacks are also classified by the norm distances used to evaluate the difference between clean and perturbed images. These include $L_0$, $L_1$, $L_2$, and $L_\infty$ norms, which measure different aspects of the perturbations \cite{huang_survey_2020}. The $L_\infty$ and $L_2$ norms are commonly used in the literature to create imperceptible perturbations, while the $L_0$ norm aims to modify only a few pixels, making attacks less visible to humans \cite{huang_survey_2020}. For instance, Papernot et al. \cite{papernot_limitations_2016} introduced the JSMA attack, which modifies specific pixels based on a saliency map that highlights the most influential pixels for classification.

Adversarial attacks are either targeted or untargeted. Targeted attacks aim to misclassify inputs into a specific class, giving more control to the attacker, while untargeted attacks simply aim to cause misclassification \cite{mekala_metamorphic_2020}. While targeted attacks are more challenging to execute, untargeted attacks are easier to implement and still highly disruptive to model performance.

\section{Method}

In this study, we systematically assess the adversarial robustness of four lightweight ConvNet architectures - MobileNet \cite{Howard2019}, MnasNet \cite{Tan2019}, SqueezeNet \cite{iandola2016}, and ShuffleNet \cite{Ma2018} - (presented in section \ref{sec:models}),  on two AHCR datasets: the Isolated Farsi Handwritten Character Dataset (IFHCDB) \cite{mozaffari2006} and the Arabic Handwritten Characters Dataset (AHCD) \cite{elsawy2017} (described in section \ref{sec:dataset}). 

\begin{figure}
    \centering    \includegraphics[width=0.8\textwidth,height=0.6\textheight,keepaspectratio]{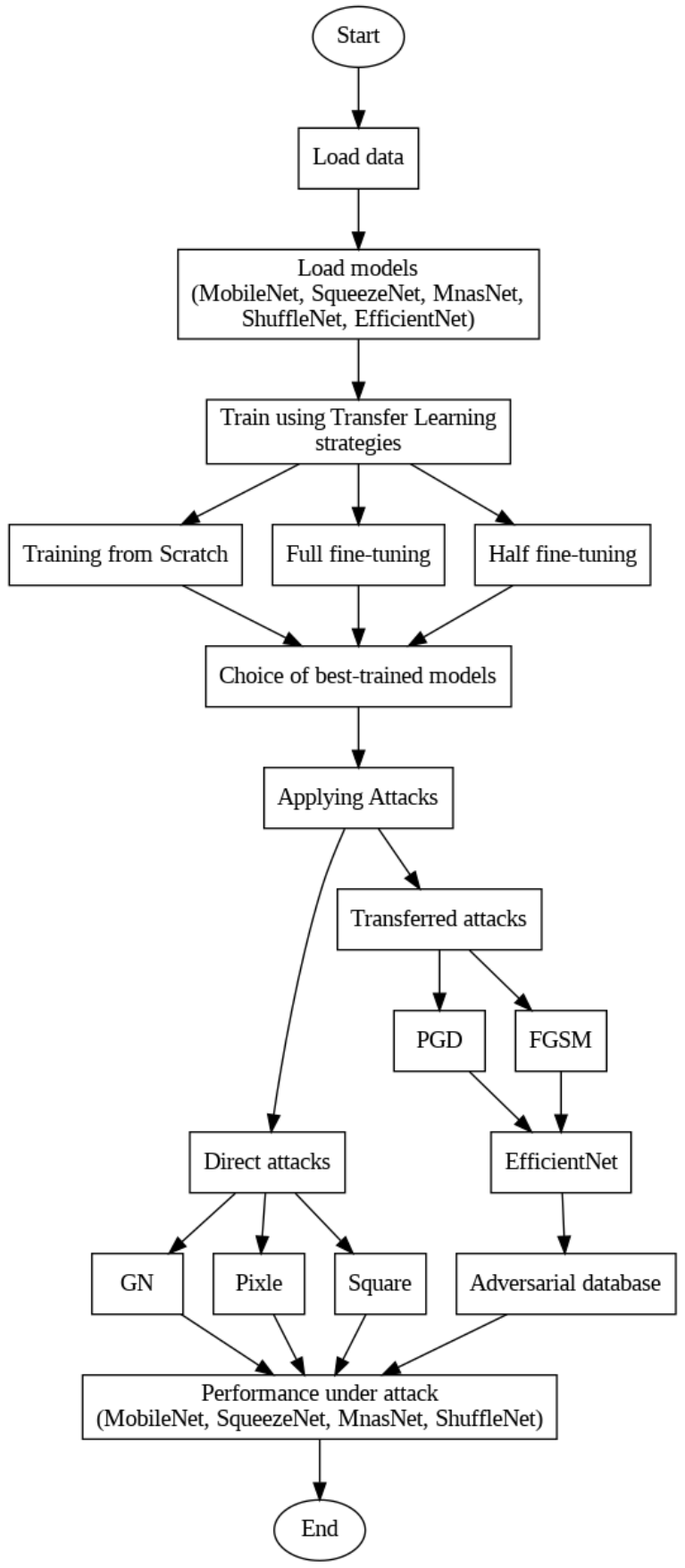}
    \caption{Flowchart of the proposed method}
    \label{fig:method}
\end{figure}

As illustrated in Fig \ref{fig:method}, the experimental pipeline starts with data loading, followed by model initialization using a pool of five candidate architectures (including EfficientNet, used solely for generating adversarial examples in transfer attack scenarios).
To ensure achieving the best performances, the models were trained using three transfer learning strategies: from scratch, full fine-tuning, and half fine-tuning. Based on the testing performance, the best-trained models were selected for adversarial testing. We evaluated model vulnerability using five black-box adversarial attacks as detailed in subsection \ref{sec:attacks}. To evaluate the attack's effectiveness, we used three key metrics: The \textbf{clean acc} representing the clean accuracy in normal conditions. The \textbf{Attack acc} represents the accuracy of the models under the attacks. The \textbf{ASR} (Attack Success Rate) measures the proportion of successfully perturbed samples that result in model misclassification.

\subsection{Attacks}
\label{sec:attacks}

Five distinct attacks were applied: Three direct black-box attacks -GN, Pixle, and Square- and two white-box attacks -PGD and FGSM. To adapt the white attacks to a black-box scenario, we first generate adversarial examples using a surrogate model, EfficientNet, and then transfer these perturbed inputs to the target models. By employing this methodology, we ensure that all attacks conform to a black-box scenario. Below is a description of the applied attacks:

\begin{itemize}

    \item \textbf {Gaussian Noise (GN)}: The GN attack introduces random noise to the input images by adding values drawn from a Gaussian distribution.
    
    \item \textbf{Square}: Square is a query-efficient black-box attack designed to work in both white-box and black-box settings. It generates adversarial examples using localized square-shaped perturbations \cite{andriushchenko_square_2020}.
    
    \item \textbf{Pixle}: Pixle manipulates specific pixels to mislead the model. This attack iteratively rearranges the pixel values, maintaining natural image statistics while achieving high attack success rates \cite{pomponi_pixle_2022}.

    \item \textbf{FGSM (Fast Gradient Sign Method)}: FGSM \cite{goodfellow_explaining_2015} is a fast white-box attack that perturbs the input image by adding noise in the direction of the model’s gradient.
    
    \item \textbf{PGD (Projected Gradient Descent)}: PGD is an iterative version of FGSM that applies multiple steps of gradient ascent to generate adversarial examples \cite{madry_towards_2019}.
\end{itemize}

\subsection{Models}
\label{sec:models}

Four models were selected for their lightweight architectures: MobileNet \cite{Howard2019}, MnasNet \cite{Tan2019}, ShuffleNet \cite{Ma2018}, and SqueezeNet \cite{iandola2016}. These models have minimal parameter counts and number of Floating Points Operations (FLOPs), making them suitable for mobile and IoT devices. They were originally designed to operate efficiently in resource-constrained environments while maintaining performance comparable to heavier models. Table \ref{tab:model_comparison} provides justification for why the studied models were chosen over other state-of-the-art models.

\begin{table}[t]
\caption{Comparison of the studied models against literature models in terms of parameters and FLOPs. Parameters in Millions}
\label{tab:model_comparison}
\centering
\begin{tabular}{|p{4cm}|p{1.2cm}|p{1cm}|}
\hline
Models & Parameters & GFLOPs \\
\hline
\multicolumn{3}{|c|}{Top performing models on ImageNet} \\
\hline
CoAtNet-7 \cite{dai_coatnet} & 2440 & 2586 \\
CoAtNet-6 \cite{dai_coatnet} & 1470 & 1521 \\
ViT-G/14 \cite{zhai_vitg} & 1843 & 2859.9 \\
DaViT-G \cite{ding_davit} & 1437 & 1038 \\
DaViT-H \cite{ding_davit} & 362 & 334 \\
\hline
\multicolumn{3}{|c|}{The models utilized in this work} \\
\hline
MobileNetV3\_small weights \cite{Howard2019} & 2.5 & 0.02 \\
ShuffleNet\_V2\_X0\_5 \cite{Ma2018} & 1.4 & 0.013 \\
MnasNet0\_5 \cite{Tan2019} & 2.2 & 0.31 \\
SqueezeNet1\_1 \cite{iandola2016} & 1.2 & 0.36 \\
\hline
\end{tabular}
\end{table}

\subsection{Datasets}
\label{sec:dataset}

Two benchmark datasets were collected for experiments: IFHCDB \cite{mozaffari2006} and AHCD \cite{elsawy2017}. AHCD consists of Arabic handwritten characters and is widely used in the literature. In contrast, IFHCDB is less known in the context of AHR as it is originally designed to support research on Farsi characters and digits. Nevertheless, it is also applicable to Arabic as it includes all Arabic characters. The characteristics of the datasets are summarized in Table \ref{tab:database}.

\begin{table*}[htbb]
\caption{Overview of dataset characteristics used in the study. Only Arabic characters from IFHCDB were retained.}
\centering
\label{tab:database}
\begin{tabular}{|c|c|c|c|c|c|c|}
\hline
Dataset & Dataset size & Train set & Test set & Images size & Content\\
\hline
IFHCDB \cite{mozaffari2006} & 70,120 &  36,017 (75\%)& 12,440 (25\%) & $77\times95$ & Character and digits \\
AHCD \cite{elsawy2017} & 16,800 & 13,440 (80\%) & 3,360 (20\%)& $32\times32$ & Characters in isolated form \\
\hline
\end{tabular}
\end{table*}

\section{Results and discussion}

This section analyses and discusses the results obtained from the experiments. Tables \ref{tab:direct_ifhcdb} and \ref{tab:direct_ahcd} and Figure \ref{fig:acc_drop} summarize the key results of the conducted experiments.

\begin{table}[htbb]
\caption{Detailed analysis of model robustness against adversarial attacks on IFHCDB}
\centering
\label{tab:direct_ifhcdb}
\begin{tabular}{|c|c|c|c|c|c|}
\hline
Attack & Model & Clean acc & Attack acc & ASR\\
\hline
GN & MobileNet & 98.08 & 28.17 & 71.49 \\
& SqueezeNet & 95.01 & 46.27 & 52.24 \\
& ShuffleNet & 97.82 & 28.99 & 70.60 \\
& MnasNet & 91.97 & 24.70 & 73.47 \\
\hline
Pixle & MobileNet & 98.08 & 0.02 & 99.98 \\
 & SqueezeNet & 95.01 & 0.00 & 100.00\\
 & ShuffleNet & 97.82 & 0.00 & 100.00 \\
 & MnasNet & 91.97 & 5.42 & 94.11 \\
\hline
Square & MobileNet & 98.08 & 23.85 & 75.67 \\
 & SqueezeNet & 95.01 & 41.81 & 56.00 \\
 & ShuffleNet & 97.82 & 24.88 & 74.57\\
 & MnasNet & 91.97 & 9.60 & 89.56 \\
\hline
PGD & MobileNet & 98.08 & 85.21 & - \\
 & SqueezeNet & 95.01 & 58.83 & - \\
 & ShuffleNet & 97.82 & 10.92 & - \\
 & MnasNet & 91.97 & 3.98 & - \\
\hline
FGSM & MobileNet & 98.08 & 85.10 & - \\
 & SqueezeNet & 95.01 & 58.09 & - \\
 & ShuffleNet & 97.82 & 10.93 & - \\
 & MnasNet & 91.97 & 3.95 &  - \\
 \hline
\end{tabular}
\end{table}

\begin{table}[htbb]
\caption{Detailed analysis of model robustness against adversarial attacks on AHCD}
\centering
\label{tab:direct_ahcd}
\begin{tabular}{|c|c|c|c|c|c|}
\hline
Attack & Model & Clean acc & Attack acc & ASR\\
\hline
GN & MobileNet & 94.91 & 3.54 & 96.27 \\
 & SqueezeNet & 95.00 & 7.65 & 92.07 \\
 & ShuffleNet & 95.15 & 3.57 & 96.28 \\
 & MnasNet & 86.13 & 9.05 & 89.73 \\
\hline
Pixle & MobileNet & 94.91 & 0.21 & 99.78 \\
 & SqueezeNet & 95.00 & 84.13 & 11.44 \\
 & ShuffleNet & 95.15 & 0.51 & 99.47 \\
 & MnasNet & 86.13 & 0.06 & 99.93 \\
\hline
Square & MobileNet & 94.91 & 3.54 & 96.27 \\
 & SqueezeNet & 95.00 & 5.69 & 94.01 \\
 & ShuffleNet & 95.15 & 3.54 & 96.28  \\
 & MnasNet & 86.13 & 2.80 & 96.75 \\
\hline
PGD & MobileNet & 94.91 & 50.01 & -  \\
 & SqueezeNet & 95.00 & 65.73  & -\\
 & ShuffleNet & 95.15 & 17.80 & - \\
 & MnasNet & 86.13 & 19.83  & -\\
\hline
FGSM & MobileNet & 94.91 & 50.19 & -\\
 & SqueezeNet & 95.00 & 65.73 & -\\
 & ShuffleNet & 95.15 & 17.80 & -\\
 & MnasNet & 86.13 & 19.62 & -\\
 \hline
\end{tabular}
\end{table}

\subsection{Attack effectiveness}

\begin{figure*}[htbp]
    \centering    \includegraphics[height=0.5\textheight,keepaspectratio]{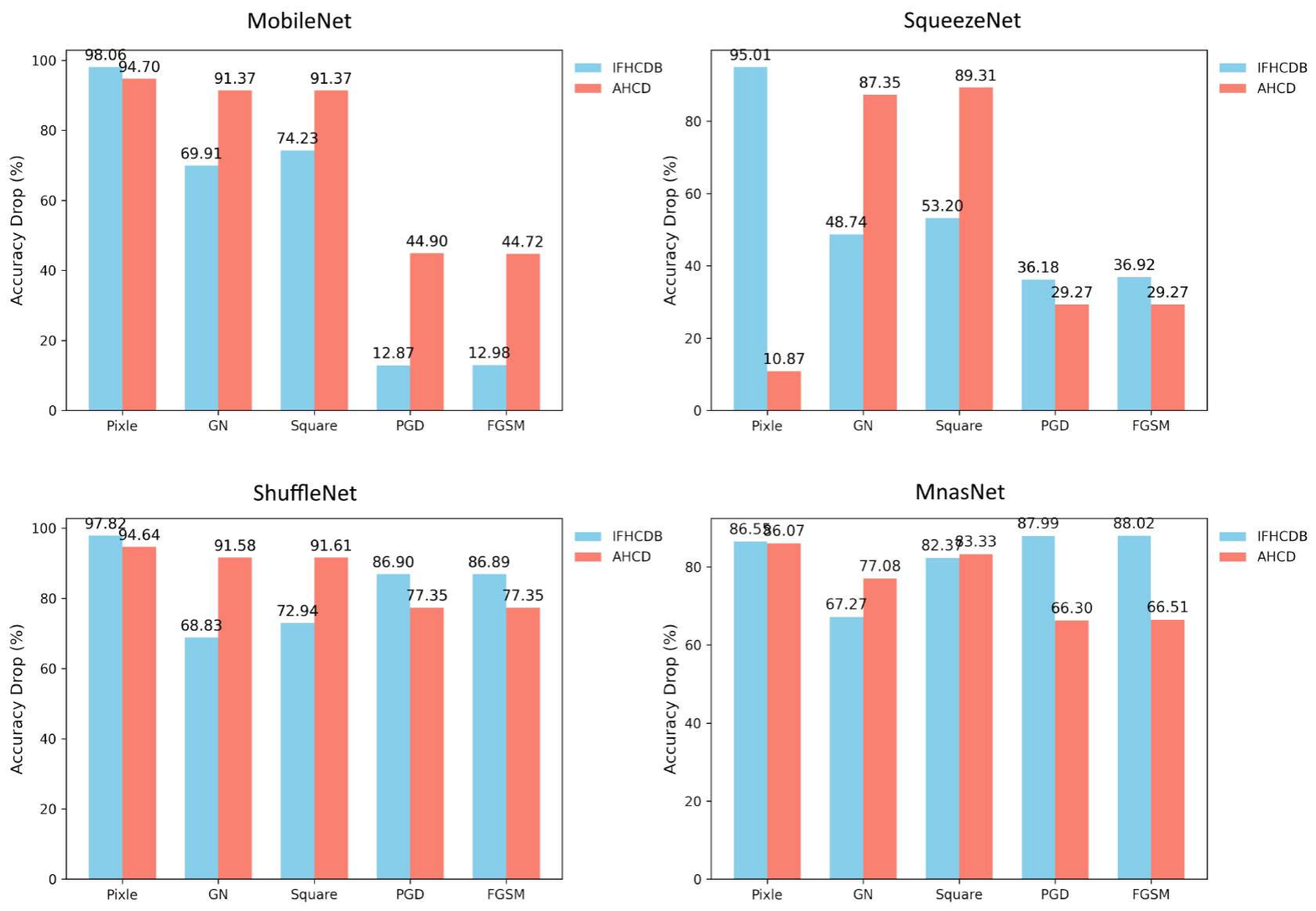}
    \caption{The accuracy drop of the studied models across attack types and datasets}
    \label{fig:acc_drop}
\end{figure*}

Based on the data presented in Tables \ref{tab:direct_ifhcdb} and \ref{tab:direct_ahcd} and figure \ref{fig:acc_drop}, the following conclusions can be drawn:

\begin{itemize}

\item \textbf{Direct attacks:} Pixle is the most successful attack across all models and datasets, reaching an ASR of 99.98\% for MobileNet and 100\% for SqueezeNet and ShuffleNet on IFHCDB. On AHCD, it acheived 99.78\% for MobileNet, and 99.93\%  ASR for MnasNet. This highlights the effectiveness of pixel-level perturbations, with one exception being SqueezeNet on AHCD, where ASR drops to 11.44\%, showing the potential resistance of this model under certain conditions. \textbf{GN} also achieves high ASRs, though less consistent, with 52.24\% on SqueezeNet and 73.47\% on MnasNet for IFHCDB, while exceeding 89\% on most AHCD models (96.27\% for MobileNet), proving models’ vulnerability to noise. \textbf{Square Attack} follows a similar trend, with 89.56\% on MnasNet and ~75\% on MobileNet for IFHCDB, and above 94\% on most AHCD models, showing slightly higher ASRs than GN across models.

\item \textbf{Transferred Attacks} vary notably across datasets and models. On IFHCDB, both attacks had a limited effect, with MobileNet retaining about 85\% accuracy. On AHCD, the attacks were more effective, reducing MobileNet’s accuracy to nearly 50\%. Overall, PGD and FGSM had a higher impact on ShuffleNet and MnasNet and a lower impact on SqueezeNet and MobileNet, especially on IFHCDB.

\end{itemize}

\subsection{Model vulnerability}

Based on the data presented in Tables \ref{tab:direct_ifhcdb} and \ref{tab:direct_ahcd} and Figure \ref{fig:acc_drop}, the following conclusions can be drawn:
\textbf{MobileNet} is highly vulnerable to Pixle attack on both datasets, with accuracy dropping to 0.02\% on IFHCDB and 0.21\% on AHCD, while showing moderate resilience to transferred attacks, maintaining about 85\% accuracy under PGD and FGSM on IFHCDB. \textbf{SqueezeNet} is extremely susceptible to Pixle attack with 0\% accuracy on IFHCDB; however, on AHCD, it resists better with 84.13\% accuracy, making it the only model to withstand the Pixle attack and suggesting better generalization on shallower or non-noisy datasets. It also shows the best resistance to GN and Square attacks on IFHCDB. \textbf{ShuffleNet} is consistently vulnerable, showing near-zero accuracy under Pixle and Square attacks on both datasets, indicating a lack of robustness regardless of attack type. \textbf{MnasNet} is highly vulnerable across all scenarios, with accuracies dropping to 5.42\% under Pixle and 9.60\% under Square attacks.

The results reveal that lightweight models are highly vulnerable to adversarial perturbations, particularly under Pixel and Square attacks, where accuracy often drops to near zero. While some resilience is observed against PGD and FGSM, it remains limited and dataset-dependent. Overall, these findings highlight the lack of robustness of compact architectures and the need for improved defense strategies.

\section{Conclusion}

This study provides a systematic examination of adversarial robustness in AHR under realistic black-box conditions, reflecting practical scenarios where attackers lack access to model architecture or parameters. The results show that ConvNet-based AHR models, though effective under normal conditions, remain highly vulnerable: Pixle attack achieves the highest Attack success rates. GN attack  even simple, achieved high attack success rates, and transferred attacks (PGD, FGSM) pose a severe real-world threat due to their zero-execution cost. Model robustness also varies, with ShuffleNet and MnasNet being most vulnerable, while MobileNet and SqueezeNet show moderate resilience. Dataset characteristics further influence outcomes, underscoring the need to consider both architecture and data properties when assessing security.

The findings raise awareness of the risks related to deep learning models' robustness in real-world AHR applications, given their growing use in sensitive domains such as finance and administration. 

Future research should focus on investigating in depth the reasons why certain models have shown significant resilience against some types of attacks and why some attacks are more aggressive than others in the context of AHR.  Another venue that must be explored is the investigation of adaptive defenses—such as adversarial training, input randomization, and denoising—to balance robustness, efficiency, and practicality.

\bibliographystyle{IEEEtran}
\bibliography{references}

@inproceedings{pomponi_pixle_2022,
	title = {Pixle: a fast and effective black-box attack based on rearranging pixels},
	shorttitle = {Pixle},
	doi = {10.1109/IJCNN55064.2022.9892966},
	booktitle = {2022 {International} {Joint} {Conference} on {Neural} {Networks} ({IJCNN})},
	author = {Pomponi, Jary and Scardapane, Simone and Uncini, Aurelio},
	month = jul,
	year = {2022},
	note = {arXiv:2202.02236 [cs, stat]},
	pages = {1--7},
}

@misc{goodfellow_explaining_2015,
	title = {Explaining and {Harnessing} {Adversarial} {Examples}},
	doi = {10.48550/arXiv.1412.6572},
	publisher = {arXiv},
	author = {Goodfellow, Ian J. and Shlens, Jonathon and Szegedy, Christian},
	month = mar,
	year = {2015},
	note = {arXiv:1412.6572 [cs, stat]},
}

@misc{madry_towards_2019,
	title = {Towards {Deep} {Learning} {Models} {Resistant} to {Adversarial} {Attacks}},
	doi = {10.48550/arXiv.1706.06083},
	publisher = {arXiv},
	author = {Madry, Aleksander and Makelov, Aleksandar and Schmidt, Ludwig and Tsipras, Dimitris and Vladu, Adrian},
	month = sep,
	year = {2019},
	note = {arXiv:1706.06083 [cs, stat]},
}

@incollection{sen_adversarial_2023,
	title = {Adversarial {Attacks} on {Image} {Classification} {Models}: {FGSM} and {Patch} {Attacks} and {Their} {Impact}},
	copyright = {https://creativecommons.org/licenses/by/3.0/legalcode},
	isbn = {978-1-83769-980-3 978-1-83768-196-9},
	shorttitle = {Adversarial {Attacks} on {Image} {Classification} {Models}},
	language = {en},
	booktitle = {Information {Security} and {Privacy} in the {Digital} {World} - {Some} {Selected} {Topics}},
	publisher = {IntechOpen},
	author = {Sen, Jaydip and Dasgupta, Subhasis},
	editor = {Sen, Jaydip and Mayer, Joceli},
	month = sep,
	year = {2023},
	doi = {10.5772/intechopen.112442},
}

@inproceedings{agrawal_impact_2021,
	address = {Orlando, FL, USA},
	title = {Impact of {Attention} on {Adversarial} {Robustness} of {Image} {Classification} {Models}},
	copyright = {https://ieeexplore.ieee.org/Xplorehelp/downloads/license-information/IEEE.html},
	isbn = {978-1-66543-902-2},
	doi = {10.1109/BigData52589.2021.9671889},
	booktitle = {2021 {IEEE} {International} {Conference} on {Big} {Data} ({Big} {Data})},
	publisher = {IEEE},
	author = {Agrawal, Prachi and Punn, Narinder Singh and Kumar Sonbhadra, Sanjay and Agarwal, Sonali},
	month = dec,
	year = {2021},
	pages = {3013--3019},
}

@inproceedings{mekala_metamorphic_2020,
	address = {Seoul Republic of Korea},
	title = {Metamorphic filtering of black-box adversarial attacks on multi-network face recognition models},
	isbn = {978-1-4503-7963-2},
	doi = {10.1145/3387940.3391483},
	language = {en},
	booktitle = {Proceedings of the {IEEE}/{ACM} 42nd {International} {Conference} on {Software} {Engineering} {Workshops}},
	publisher = {ACM},
	author = {Mekala, Rohan Reddy and Porter, Adam and Lindvall, Mikael},
	month = jun,
	year = {2020},
	pages = {410--417},
}

@article{el_khayati_cnn-based_2024,
	title = {{CNN}-based {Methods} for {Offline} {Arabic} {Handwriting} {Recognition}: {A} {Review}},
	volume = {56},
	issn = {1573-773X},
	shorttitle = {{CNN}-based {Methods} for {Offline} {Arabic} {Handwriting} {Recognition}},
	doi = {10.1007/s11063-024-11544-w},
	language = {en},
	number = {2},
	journal = {Neural Processing Letters},
	author = {El Khayati, Mohsine and Kich, Ismail and Taouil, Youssef},
	month = mar,
	year = {2024},
	pages = {115},
}

@inproceedings{Tan2019,
  title = {MnasNet: Platform-Aware Neural Architecture Search for Mobile},
  DOI = {10.1109/cvpr.2019.00293},
  booktitle = {2019 IEEE/CVF Conference on Computer Vision and Pattern Recognition (CVPR)},
  publisher = {IEEE},
  author = {Tan,  Mingxing and Chen,  Bo and Pang,  Ruoming and Vasudevan,  Vijay and Sandler,  Mark and Howard,  Andrew and Le,  Quoc V.},
  year = {2019},
  month = jun,
  pages = {2815–2823}
}

@article{rabi_convolutional_2024,
	title = {Convolutional {Arabic} handwriting recognition system based {BLSTM}-{CTC} using {WBS} decoder},
	volume = {4},
	copyright = {https://creativecommons.org/licenses/by/4.0},
	issn = {2809-7467, 2809-7599},
	doi = {10.47679/ijasca.v3i2.52},
	number = {1},
	journal = {International Journal of Advanced Science and Computer Applications},
	author = {Rabi, Mouhcine},
	month = jan,
	year = {2024},
}

@inproceedings{korichi_arabic_2020,
	address = {Giza, Egypt},
	title = {Arabic handwriting recognition: {Between} handcrafted methods and deep learning techniques},
	copyright = {https://ieeexplore.ieee.org/Xplorehelp/downloads/license-information/IEEE.html},
	isbn = {978-1-72818-855-3},
	shorttitle = {Arabic handwriting recognition},
	doi = {10.1109/ACIT50332.2020.9300121},
	booktitle = {2020 21st {International} {Arab} {Conference} on {Information} {Technology} ({ACIT})},
	publisher = {IEEE},
	author = {Korichi, Aicha and Slatnia, Sihem and Aiadi, Oussama and Tagougui, Najiba and Kherallah, Monji},
	month = nov,
	year = {2020},
	pages = {1--6},
}

@inproceedings{shtaiwi_end--end_2022,
	address = {Irbid, Jordan},
	title = {End-to-{End} {Machine} {Learning} {Solution} for {Recognizing} {Handwritten} {Arabic} {Documents}},
	copyright = {https://doi.org/10.15223/policy-029},
	isbn = {978-1-66548-097-0},
	doi = {10.1109/ICICS55353.2022.9811155},
	booktitle = {2022 13th {International} {Conference} on {Information} and {Communication} {Systems} ({ICICS})},
	publisher = {IEEE},
	author = {Shtaiwi, Reem E. and Abandah, Gheith A. and Sawalhah, Safaa A.},
	month = jun,
	year = {2022},
	pages = {180--185},
}

@article{elleuch_novel_2016,
	title = {A novel architecture of {CNN} based on {SVM} classifier for recognising {Arabic} handwritten script},
	volume = {15},
	issn = {1740-8865, 1740-8873},
	doi = {10.1504/IJISTA.2016.080103},
	language = {en},
	number = {4},
	journal = {International Journal of Intelligent Systems Technologies and Applications},
	author = {Elleuch, Mohamed and Tagougui, Najiba and Kherallah, Monji},
	year = {2016},
	pages = {323},
}

@article{ahmed_novel_2021,
	title = {Novel {Deep} {Convolutional} {Neural} {Network}-{Based} {Contextual} {Recognition} of {Arabic} {Handwritten} {Scripts}},
	volume = {23},
	copyright = {https://creativecommons.org/licenses/by/4.0/},
	issn = {1099-4300},
	doi = {10.3390/e23030340},
	language = {en},
	number = {3},
	journal = {Entropy},
	author = {Ahmed, Rami and Gogate, Mandar and Tahir, Ahsen and Dashtipour, Kia and Al-tamimi, Bassam and Hawalah, Ahmad and El-Affendi, Mohammed A. and Hussain, Amir},
	month = mar,
	year = {2021},
	pages = {340},
}

@incollection{chen_fawa_2021,
	address = {Cham},
	title = {{FAWA}: {Fast} {Adversarial} {Watermark} {Attack} on {Optical} {Character} {Recognition} ({OCR}) {Systems}},
	volume = {12459},
	isbn = {978-3-030-67663-6 978-3-030-67664-3},
	shorttitle = {{FAWA}},
	language = {en},
	booktitle = {Machine {Learning} and {Knowledge} {Discovery} in {Databases}},
	publisher = {Springer International Publishing},
	author = {Chen, Lu and Sun, Jiao and Xu, Wei},
	editor = {Hutter, Frank and Kersting, Kristian and Lijffijt, Jefrey and Valera, Isabel},
	year = {2021},
	doi = {10.1007/978-3-030-67664-3_33},
	note = {Series Title: Lecture Notes in Computer Science},
	pages = {547--563},
}

@misc{iandola2016,
  doi = {10.48550/ARXIV.1602.07360},
  author = {Iandola,  Forrest N. and Han,  Song and Moskewicz,  Matthew W. and Ashraf,  Khalid and Dally,  William J. and Keutzer,  Kurt},
  title = {SqueezeNet: AlexNet-level accuracy with 50x fewer parameters and \&lt;0.5MB model size},
  publisher = {arXiv},
  year = {2016},
  copyright = {arXiv.org perpetual,  non-exclusive license}
}

@inbook{Ma2018,
  title = {ShuffleNet V2: Practical Guidelines for Efficient CNN Architecture Design},
  ISBN = {9783030012649},
  ISSN = {1611-3349},
  DOI = {10.1007/978-3-030-01264-9_8},
  booktitle = {Computer Vision – ECCV 2018},
  publisher = {Springer International Publishing},
  author = {Ma,  Ningning and Zhang,  Xiangyu and Zheng,  Hai-Tao and Sun,  Jian},
  year = {2018},
  pages = {122–138}
}

@inproceedings{mozaffari2006,
  author    = {Saeed Mozaffari and Karim Faez and Farshad Faradji and Majid Ziaratban and S. M. Golzan},
  title     = {A Comprehensive Isolated Farsi/Arabic Character Database for Handwritten OCR Research},
  booktitle = {Proceedings of the 10th International Workshop on Frontiers in Handwriting Recognition},
  pages     = {385--389},
  year      = {2006},
  address   = {La Baule, France}
}

@article{elsawy2017,
  author  = {Ahmed Elsawy and Mohamed Loey and Hazem El-Bakry},
  title   = {Arabic Handwritten Characters Recognition using Convolutional Neural Network},
  journal = {WSEAS Transactions on Computer Research},
  volume  = {5},
  pages   = {11--19},
  year    = {2017}
}

@article{akhtar_threat_2018,
	title = {Threat of {Adversarial} {Attacks} on {Deep} {Learning} in {Computer} {Vision}: {A} {Survey}},
	volume = {6},
	copyright = {https://ieeexplore.ieee.org/Xplorehelp/downloads/license-information/OAPA.html},
	issn = {2169-3536},
	shorttitle = {Threat of {Adversarial} {Attacks} on {Deep} {Learning} in {Computer} {Vision}},
	doi = {10.1109/ACCESS.2018.2807385},
	journal = {IEEE Access},
	author = {Akhtar, Naveed and Mian, Ajmal},
	year = {2018},
	pages = {14410--14430},
}

@article{huang_survey_2020,
	title = {A survey of safety and trustworthiness of deep neural networks: {Verification}, testing, adversarial attack and defence, and interpretability},
	volume = {37},
	issn = {15740137},
	shorttitle = {A survey of safety and trustworthiness of deep neural networks},
	doi = {10.1016/j.cosrev.2020.100270},
	language = {en},
	journal = {Computer Science Review},
	author = {Huang, Xiaowei and Kroening, Daniel and Ruan, Wenjie and Sharp, James and Sun, Youcheng and Thamo, Emese and Wu, Min and Yi, Xinping},
	month = aug,
	year = {2020},
	pages = {100270},
}

@inproceedings{castro_improvement_2021,
	address = {Cagliari, Italy},
	title = {Improvement {Optical} {Character} {Recognition} for {Structured} {Documents} using {Generative} {Adversarial} {Networks}},
	copyright = {https://ieeexplore.ieee.org/Xplorehelp/downloads/license-information/IEEE.html},
	isbn = {978-1-66545-843-6},
	doi = {10.1109/ICCSA54496.2021.00046},
	booktitle = {2021 21st {International} {Conference} on {Computational} {Science} and {Its} {Applications} ({ICCSA})},
	publisher = {IEEE},
	author = {Castro, Jose D. Bermudez and Canchumuni, Smith W. Arauco and Villalobos, Cristian E. Munoz and Cordeiro, Fabio Correa and Alexandre, Antonio Marcelo Azevedo and Pacheco, Marco A. Cavalcanti},
	month = sep,
	year = {2021},
	pages = {285--292},
}

@article{devan_one-word_2020,
	title = {One-{Word} {Answer} {Correction} using {Deep} {Learning} {Models} and {OCR}},
	volume = {9},
	issn = {22773878},
	doi = {10.35940/ijrte.B3849.079220},
	number = {2},
	journal = {International Journal of Recent Technology and Engineering (IJRTE)},
	author = {Devan, K. P. K. and Sruthi Prabakaran and S Tamizhazhagan and S Vaishnavi},
	month = jul,
	year = {2020},
	pages = {679--682},
}

@article{mohd_quranic_2021,
	title = {Quranic {Optical} {Text} {Recognition} {Using} {Deep} {Learning} {Models}},
	volume = {9},
	copyright = {https://creativecommons.org/licenses/by/4.0/legalcode},
	issn = {2169-3536},
	doi = {10.1109/ACCESS.2021.3064019},
	journal = {IEEE Access},
	author = {Mohd, Masnizah and Qamar, Faizan and Al-Sheikh, Idris and Salah, Ramzi},
	year = {2021},
	pages = {38318--38330},
}

@misc{szegedy_intriguing_2014,
	title = {Intriguing properties of neural networks},
	doi = {10.48550/arXiv.1312.6199},
	publisher = {arXiv},
	author = {Szegedy, Christian and Zaremba, Wojciech and Sutskever, Ilya and Bruna, Joan and Erhan, Dumitru and Goodfellow, Ian and Fergus, Rob},
	month = feb,
	year = {2014},
	note = {arXiv:1312.6199},
}

@inproceedings{irfan_towards_2021,
	address = {Islamabad, Pakistan},
	title = {Towards {Deep} {Learning}: {A} {Review} {On} {Adversarial} {Attacks}},
	copyright = {https://ieeexplore.ieee.org/Xplorehelp/downloads/license-information/IEEE.html},
	isbn = {978-1-66543-293-1},
	shorttitle = {Towards {Deep} {Learning}},
	doi = {10.1109/ICAI52203.2021.9445247},
	booktitle = {2021 {International} {Conference} on {Artificial} {Intelligence} ({ICAI})},
	publisher = {IEEE},
	author = {Irfan, Muhammad Maaz and Ali, Sheraz and Yaqoob, Irfan and Zafar, Numan},
	month = apr,
	year = {2021},
	pages = {91--96},
}

@misc{huang_adversarial_2017,
	title = {Adversarial {Attacks} on {Neural} {Network} {Policies}},
	publisher = {arXiv},
	author = {Huang, Sandy and Papernot, Nicolas and Goodfellow, Ian and Duan, Yan and Abbeel, Pieter},
	month = feb,
	year = {2017},
	note = {arXiv:1702.02284 [cs]},
}

@inproceedings{papernot_limitations_2016,
	title = {The {Limitations} of {Deep} {Learning} in {Adversarial} {Settings}},
	doi = {10.1109/EuroSP.2016.36},
	booktitle = {2016 {IEEE} {European} {Symposium} on {Security} and {Privacy} ({EuroS}\&{P})},
	author = {Papernot, Nicolas and McDaniel, Patrick and Jha, Somesh and Fredrikson, Matt and Celik, Z. Berkay and Swami, Ananthram},
	month = mar,
	year = {2016},
	pages = {372--387},
}

@inproceedings{Howard2019,
  title = {Searching for MobileNetV3},
  DOI = {10.1109/iccv.2019.00140},
  booktitle = {2019 IEEE/CVF International Conference on Computer Vision (ICCV)},
  publisher = {IEEE},
  author = {Howard,  Andrew and Sandler,  Mark and Chen,  Bo and Wang,  Weijun and Chen,  Liang-Chieh and Tan,  Mingxing and Chu,  Grace and Vasudevan,  Vijay and Zhu,  Yukun and Pang,  Ruoming and Adam,  Hartwig and Le,  Quoc},
  year = {2019},
  month = oct,
  pages = {1314–1324}
}

@article{tagougui_online_2012,
	title = {Online {Arabic} handwriting recognition: a survey},
	volume = {16},
	shorttitle = {Online {Arabic} handwriting recognition},
	doi = {10.1007/s10032-012-0186-8},
	language = {en},
	journal = {International Journal on Document Analysis and Recognition (IJDAR)},
	author = {Tagougui, Najiba and Alimi, Adel M. and Kherallah, Monji},
	month = may,
	year = {2012},
}

@inproceedings{andriushchenko_square_2020,
	address = {Cham},
	title = {Square {Attack}: {A} {Query}-{Efficient} {Black}-{Box} {Adversarial} {Attack} via {Random} {Search}},
	isbn = {978-3-030-58592-1},
	shorttitle = {Square {Attack}},
	doi = {10.1007/978-3-030-58592-1_29},
	language = {en},
	booktitle = {Computer {Vision} – {ECCV} 2020},
	publisher = {Springer International Publishing},
	author = {Andriushchenko, Maksym and Croce, Francesco and Flammarion, Nicolas and Hein, Matthias},
	editor = {Vedaldi, Andrea and Bischof, Horst and Brox, Thomas and Frahm, Jan-Michael},
	year = {2020},
	pages = {484--501},
}

@misc{dai_coatnet,
  author = {Z. Dai and H. Liu and Q. V. Le and M. Tan},
  title  = {CoAtNet: Marrying Convolution and Attention for All Data Sizes},
  year   = {2021},
  eprint = {2106.04803},
  archivePrefix = {arXiv},
  primaryClass = {cs}
}

@inproceedings{ding_davit,
  author    = {M. Ding and B. Xiao and N. Codella and P. Luo and J. Wang and L. Yuan},
  title     = {DaViT: Dual Attention Vision Transformers},
  year      = {2022},
  booktitle = {Proceedings of the Conference},
  doi       = {10.48550/arXiv.2204.03645}
}

@article{zhai_vitg,
  author = {Zhai, X. and Kolesnikov, A. and Houlsby, N. and Beyer, L.},
  title  = {Scaling Vision Transformers},
  journal = {arXiv:2106.04560 [cs]},
  year   = {2021},
}

@article{ElKhayati2025_laveraging,
  title = {Leveraging Transfer Learning and Mobile-Enabled Convolutional Neural Networks for Improved Arabic Handwritten Character Recognition},
  volume = {13},
  ISSN = {2169-3536},
  DOI = {10.1109/access.2025.3613265},
  journal = {IEEE Access},
  publisher = {Institute of Electrical and Electronics Engineers (IEEE)},
  author = {El Khayati,  Mohsine and Maafiri,  Ayyad and Himeur,  Yassine and Ali Alkhazaleh,  Hamzah and Atalla,  Shadi and Mansoor,  Wathiq},
  year = {2025},
  pages = {166104–166126}
}

\end{document}